\pdfoutput=1

\documentclass[11pt]{article}

\usepackage[final]{acl}

\usepackage{times}
\usepackage{latexsym}
\usepackage[T1]{fontenc}

\usepackage[utf8]{inputenc}

\usepackage{microtype}

\usepackage{inconsolata}

\usepackage{graphicx}

\usepackage{hyperref}
\usepackage{url}
\usepackage{booktabs}
\usepackage{pifont}
\usepackage{booktabs}
\usepackage{multirow}
\usepackage{makecell}
\usepackage[fixed]{fontawesome5}
\usepackage{wrapfig}
\usepackage{marvosym}
\usepackage[linesnumbered,ruled,vlined]{algorithm2e}

\definecolor{darkblue}{rgb}{0, 0, 0.5}

\hypersetup{colorlinks=true, citecolor=darkblue, linkcolor=darkblue, urlcolor=darkblue}
\usepackage{colortbl}

\definecolor{c1}{HTML}{217074}
\definecolor{c2}{HTML}{37745B}
\definecolor{c3}{HTML}{8B9D77}
\definecolor{c4}{HTML}{E7EAEF}
\definecolor{c5}{HTML}{EDC5AB}
\definecolor{smurf}{HTML}{34629E}

\usepackage[most]{tcolorbox}
\usepackage{float}
\usepackage{xspace}
\tcbset{
  aibox/.style={
    width=474.18663pt,
    top=10pt,
    colback=white,
    colframe=black,
    colbacktitle=black,
    enhanced,
    center,
    attach boxed title to top left={yshift=-0.1in,xshift=0.15in},
    boxed title style={boxrule=0pt,colframe=white,},
  }
}
\newtcolorbox{AIbox}[2][]{aibox,title=#2,#1}

\definecolor{my_green}{RGB}{51,102,0}
\definecolor{my_red}{RGB}{204, 0, 0}
\newcommand{\colorcmark}{\textcolor{my_green}{\ding{52}}}
\newcommand{\colorxmark}{\textcolor{my_red}{\ding{55}}}
\title{Smurfs: Multi-Agent System using Context-Efficient DFSDT for Tool Planning}
\author{Junzhi Chen\footnotemark[1], Juhao Liang\footnotemark[1], Benyou Wang\textsuperscript{\Letter} 
\\
The Chinese University of Hong Kong, Shenzhen\\
Shenzhen Research Institute of Big Data\\
\href{mailto:wangbenyou@cuhk.edu.cn}{wangbenyou@cuhk.edu.cn}
}

\begin{document}
\maketitle
\renewcommand{\thefootnote}{\fnsymbol{footnote}} 
\footnotetext[1]{These authors contributed equally to this work.} 
\footnotetext{\Letter Corresponding author.} 

\begin{abstract}


Teaching large language models (LLMs) to use tools for solving complex problems can grant them human-like reasoning abilities. ReAct and its variants are popular frameworks for tool use in both single-agent and multi-agent systems. To address issues like error propagation and limited exploration in ReAct, the Deep First Search Decision Tree (DFSDT) was proposed, but it faces challenges such as rollback instability, redundant context, and premature termination in single-agent settings. We introduce "Smurfs," a novel multi-agent system (MAS) that enhances DFSDT with a modular, context-efficient, and training-free design. Smurfs surpasses baseline methods in both the open-ended StableToolBench and the closed-ended HotpotQA tasks, reducing token usage by 60.9\% compared to DFSDT and enabling Mistral-7b to perform on par with GPT-4-DFSDT. Extensive ablation studies confirm the effectiveness of Smurfs' core components, offering valuable insights for the construction and interpretation of MAS, and paving the way for future exploration. We release the code at \href{https://github.com/FreedomIntelligence/Smurfs}{https://github.com/FreedomIntelligence/Smurfs}.

\end{abstract}

\begin{figure*}[htb]
    \centering
    \includegraphics[width=1.0\textwidth]{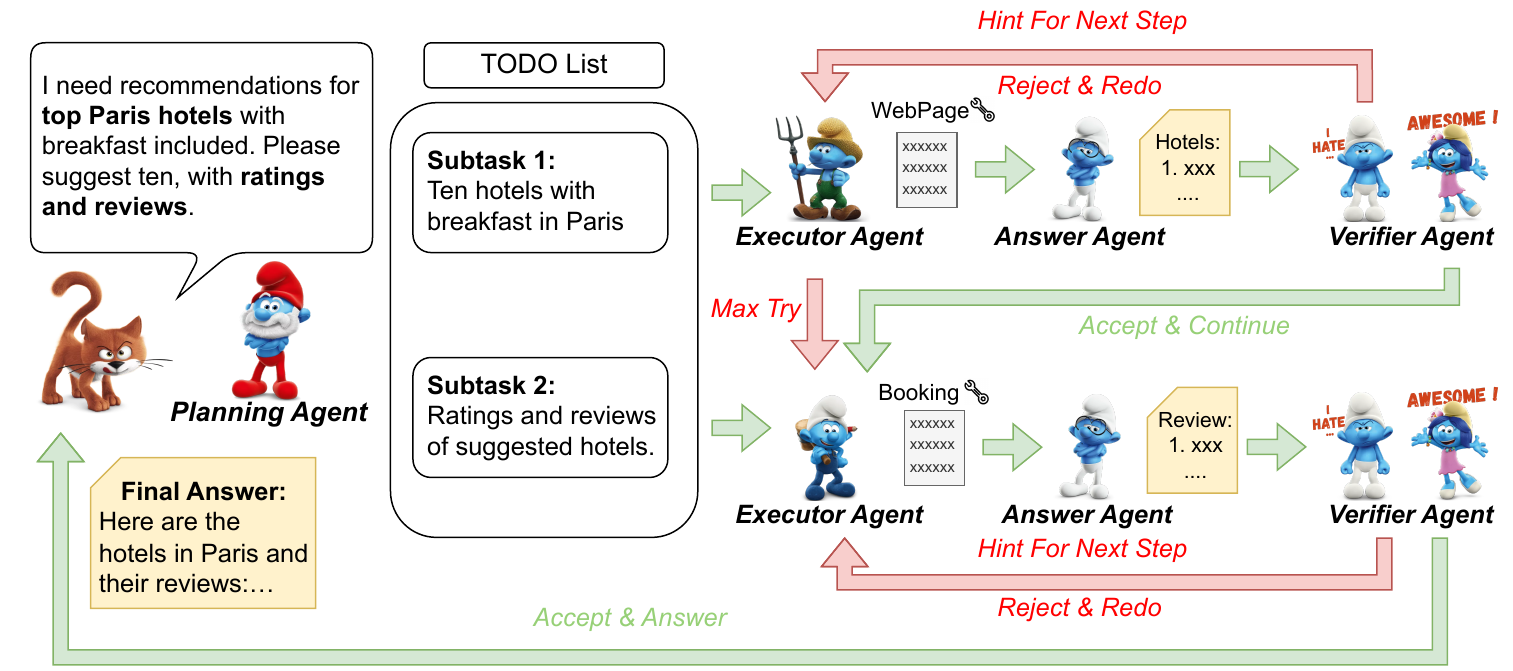}
    \caption{Demonstration of the whole process of the Smurfs framework. }
    \label{fig:multi-agent}
\end{figure*}

\section{Introduction}
\label{sec: intro}


The ability to manipulate tools for complex tasks has long been considered a distinctive characteristic of humans~\cite{oakley1972man, ambrose2001paleolithic}.  Can we extend this ability to today’s large language models (LLMs), enabling them to utilize multiple tools to perform complex tasks beyond their inherent capabilities? If LLMs can use external tools to access knowledge or execute tasks beyond their fixed language modeling capabilities, we can shift the focus of LLM training towards enhancing their reasoning and tool-use skills. This shift would allow tools to supplement what LLMs should know or execute, thereby improving the parameter efficiency of the LLMs.

\begin{table}[htb]
    \centering
    \footnotesize
    \scalebox{0.7}{
        \begin{tabular}{l|cccc}
            \toprule
                 & \makecell{Pass Rate $\uparrow$\\(\%)} & \makecell{Win Rate $\uparrow$\\(\%)} & \makecell{\# of Tokens\\per request $\downarrow$} & \makecell{\# of Tokens\\per query $\downarrow$} \\
                 \midrule
                 ReAct & $44.4_{\pm 1.1}$ & \textit{base} & 1,424 & 6,479 \\
                 DFSDT & $55.4_{\pm 2.0}$ & 60.4 & 1,743 & 20,714 \\
                 \textbf{Smurfs (ours)} & $57.4_{\pm 1.1}$ & 62.4 & 459 & 8,096 \\
            \bottomrule
        \end{tabular}
    }
    \caption{Comparison of token cost and performance between tool planning methods over StableToolBench. Existing methods, \textit{ReAct} and \textit{DFSDT}, have limitations due to high token costs or poor performance. The results are averaged over the subtasks within StableToolBench.}
    \label{tab:token cost}
\end{table}

In previous multi-agent systems for tool planning, methods like Chain-of-Thought (CoT)~\cite{wei2023chainofthought}, ReAct~\cite{yao2022react}, and the more advanced DFSDT~\cite{qin2024toolllm} have been proposed to enhance LLMs' ability to handle complex multi-step tasks. However, these approaches face notable limitations. ReAct has trouble in handling \textit{error propagation} and \textit{limited exploration}. DFSDT incorporates a rollback mechanism and depth-first search approach to address limitations of ReAct, but it suffers from \textit{instability} when the base model struggles with long-context reasoning. It also introduces inefficiencies due to \textit{redundant context} handling and risks \textit{premature termination} when solving multi-step problems. These challenges highlight the need for further innovation to better manage context and reasoning complexity in multi-tool planning systems.

In this paper, we introduce \textbf{\textit{`Smurfs'}}, an innovative multi-agent system (MAS) framework inspired by the collaborative and versatile nature of its namesake cartoon characters. The proposed framework leverages enhanced DFSDT to perform complex tool planning tasks without the need for additional training. The effectiveness of Smurfs is demonstrated through both open-ended and closed-ended tool planning benchmark experiments~\cite{guo2024stabletoolbench, yang2018hotpotqa}, where it consistently outperforms baseline methods. An ablation study, followed by a case study, further investigates the reasons behind this effectiveness. These results establish a new state-of-the-art in the field and provide concrete evidence of the advantages of a multi-agent approach in enhancing LLM capabilities.

The contributions of this paper can be summarized as follows:
\begin{enumerate}
    \item We introduce a highly modular, context-efficient, and training-free MAS framework that utilizes an enhanced DFSDT to improve the tool planning capabilities of LLMs. Experiments demonstrate the effectiveness of this approach, which also proves to be more cost-efficient compared to existing methods.
    \item Through ablation studies, we uncover the underlying factors contributing to the effectiveness of the MAS framework, offering valuable insights for future research. 
\end{enumerate}

\section{Related Work}

To augment LLMs to do multi-tool planning for solving complex problems, previous work has seen numerous attempts. Chain-of-Thought ~\cite{wei2023chainofthought} is the first to propose the method of thought and answer chain reasoning. ReAct~\cite{yao2022react} further introduce the thought-action-observation format for tool chain reasoning, leading to the development of various multi-tool planning methods~\cite{chen2023fireact, xu2023rewoo, shinn2023reflexion}. The latest work, Deep First Search Decision Tree (DFSDT)~\cite{qin2024toolllm}, is proposed to address the inherent limitations of CoT and ReACT: 
\textbf{(1) Error Propagation: } Error occurs at early stage of planning will result in wrong answer in the end, but it can only be identified until reaching the end of the planning chain
\textbf{(2) Limited Exploration}: Single solution chain can't explore the planning space completely.
\\
\\
DFSDT is powerful in addressing multi-tool planning problems. Its core concept involves employing a depth-first search (DFS) approach for multi-tool planning and backtracks whenever an LLM think the solving process has entered a wrong state (for more details, see Appendix~\ref{appendix: DFSDT Details}). When a tool fails or is deemed inadequate for solving the current problem, DFSDT backtracks to the previous solution state and attempts to resolve the issue using another solution plan. However, three limitations are identified with the mechanism of DFSDT: 

\paragraph{Limit I: Instability of the Rollback Mechanism}
The rollback mechanism in DFSDT is determined by the model. The number of steps to roll back and the selection of new tools after rollback are guided using prompt containing the errors encountered in the previous failed trajectory. When the model is sufficiently robust, this rollback mechanism serves as a highly flexible and efficient planning strategy. However, when the model's capability is insufficient, it will fail to execute the correct rollback mechanism, i.e. retry the same error tools or roll back too far.

\paragraph{Limit II: Redundant Context}
In the process of planning with DFSDT, each tool plan is generated using the entire conversation history (including all the thoughts, actions, action inputs and tool responses) as context. However, in reality, each step of tool planning only requires a very small portion of the relevant history for effective planning. 

The context redundancy not only increases computational overhead but also reduces the accuracy of model inference due to the inclusion of irrelevant historical data. As highlighted by~\cite{liu2024lost}, redundant context become particularly noticeable in tasks requiring assimilation and processing of large inputs, like verbose tool documents and API responses. The situation worsens when LLMs are supplemented with external information, such as document retrieval or online searching~\cite{petroni2020context, ram2023context, mallen2022not}. Although numerous language models capable of handling larger contexts are emerging~\cite{dai2019transformer, dao2022flashattention}, they often face significant performance degradation when the important information is located at some positions~\cite{liu2024lost, shi2023large}, which is known as the `\textit{lost-in-the-middle}' problem.

\paragraph{Limit III: Premature Termination}
The termination mechanism set by DFSDT involves adding a termination tool to the model's selectable toolkit. When the model selects this termination tool, DFSDT stops and provides an answer. However, in practical applications, this mechanism often prematurely terminates when dealing with complex problems requiring multi-step reasoning. We hypothesize that this issue arises due to the redundant interference of other tool information and history information, which disrupts the model's ability to judge whether the original problem should be terminated. Instead, the model focuses on whether the current sub-problem requires termination, leading the mechanism to terminate after resolving the sub-problem.

In conclusion, DFSDT relies highly on the base model's reasoning ability, especially long context reasoning ability to make roll back decision, termination decision and tool choice decision at the same time, which is a very difficult task even with the most powerful LLM like GPT4~\cite{yuan2024easytool}.

\paragraph{Multi-Agent System}
To address the limitations inherent in DFSDT and to further enhance LLM's multi-tool planning capabilities, MAS has emerged as a natural solution. Inspired by human social division of labor and cooperation, MAS aim to enable AI agents to accomplish more complex tasks through the division of labor and collaboration. By decomposing the task of DFSDT to multiple agents and giving them only the information they need, we can enable LLMs to use DFSDT more effectively and more efficiently.

\begin{table}[h!]
\centering
\footnotesize
\scalebox{0.52}{
\begin{tabular}{lccccc}
\hline
\toprule
\textbf{Method} & \textbf{Multi-Agent} & \textbf{Training} & \textbf{Generality} & \textbf{Reflection} & \textbf{Planning} \\
\Xhline{1px}
\textsc{ReAct} \cite{yao2022react} & \colorxmark & \colorxmark & \colorcmark & \colorxmark & Iterative\\
Reflexion \cite{shinn2023reflexion} & \colorxmark & \colorxmark & \colorcmark & \colorcmark & Iterative \\
Chameleon \cite{lu2023chameleon} & \colorxmark & \colorxmark & \colorcmark & \colorxmark & Global\\
HuggingGPT \cite{shen2023hugginggpt} & \colorxmark & \colorxmark & \colorcmark & \colorxmark & Global \\
BOLAA \cite{liu2023bolaa} & \colorcmark & \colorxmark & \colorcmark & \colorxmark & Iterative \\
AgentVerse \cite{chen2023agentverse} & \colorcmark & \colorxmark & \colorcmark & \colorxmark & Iterative \\
\textsc{FireAct} \cite{chen2023fireact} & \colorxmark & \colorcmark & \colorxmark & \colorcmark & Iterative \\
\textsc{DFSDT} \cite{qin2024toolllm} & \colorxmark & \colorcmark & \colorxmark & \colorxmark & Iterative \\
\textsc{RestGPT} \cite{song2023restgpt} & \colorcmark & \colorxmark & \colorcmark & \colorxmark & Iterative \\
Lumos \cite{yin2024agent} & \colorcmark & \colorcmark & \colorxmark & \colorxmark & Iterative or Global \\
AutoAct \cite{qiao2024autoact} & \colorcmark & \colorcmark & \colorxmark & \colorcmark & Iterative \\
\textbf{Smurfs} (Ours) & \colorcmark & \colorxmark & \colorcmark & \colorcmark & Iterative and Global \\
\bottomrule
\hline
\end{tabular}
}
\caption{Comparison of tool use systems.}
\label{tab:compare}
\end{table}

Previous work on multi-agent system mainly focus on coding and society simulation area~\cite{hong2023metagpt, qian2024chatdevcommunicativeagentssoftware, park2023generativeagentsinteractivesimulacra, li2023camelcommunicativeagentsmind, wu2023autogenenablingnextgenllm}. For tool-use scenario, most multi-agent systems still use the ReAct style reasoning~\cite{qiao2024autoact, chen2023agentverse, yuan2024easytool, liu2023bolaa, song2023restgpt, yin2024agent, xu2023rewoo}, only using multi-agent discussion and revision to increase reasoning quality, which still inherits the limitation of ReAct. 
Therefore, this paper aims to construct a novel MAS framework to address the aforementioned limitations.
Table~\ref{tab:compare} shows the difference between different tool-use systems. More detailed discussion can be seen in Appendix~\ref{appendix: Comparison}. 

\paragraph{Token Compression}
Token compression refers to compressing tokens fed into LLMs while preserving inference performance, thus reducing computational overhead and mitigating the constraints imposed by long context limits. Previous works~\cite{mu2024learningcompresspromptsgist, fu2024camphorcollaborativeagentsmultiinput} have explored token compression techniques, with a focus on compressing token embeddings. In contrast, Smurfs achieves context compression by filtering the input context for each tool planning process. Table~\ref{tab:token compression} shows the detailed difference between different compression techniques.

In the future, these compression methods can be used together in multi-tool planning scenarios to achieve more efficient token compression. Each agent's system prompt could be compressed using gist tokens~\cite{mu2024learningcompresspromptsgist}, tool descriptions could be managed with CAMPHOR's compression approach~\cite{fu2024camphorcollaborativeagentsmultiinput}, and the context input by each agent could be compressed using Smurfs.

\begin{figure*}[htb]
    \centering
    \includegraphics[width=0.85\textwidth]{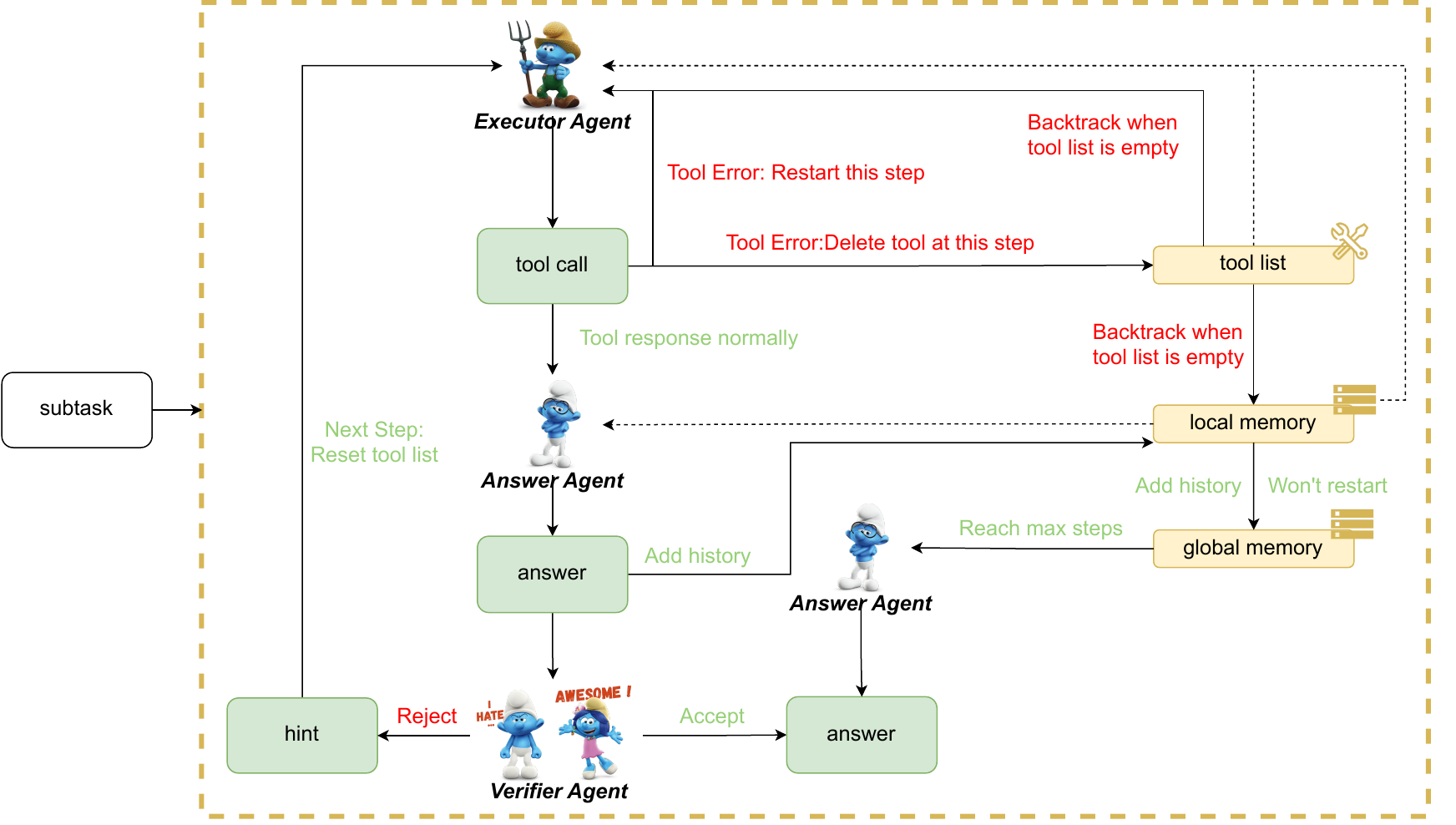}
    \caption{Details of the subtask-solving process of the Smurfs framework. The dotted line represents that the agent can see the memory and the full line stands for operation.}
    \label{fig:multi-agent_detail1}
\end{figure*}

\section{Smurfs: MAS with Context Efficient DFSDT}
\label{sec: methodology}

\textcolor{smurf}{\textit{\textbf{\large The Smurfs}, the beloved cartoon characters, symbolize unity and resourcefulness, and are good at using tools to overcome any challenge they encounter. }}

\subsection{Framework Overview}

Figure~\ref{fig:multi-agent} illustrates the entire workflow for the Smurfs framework. Initially, the \textbf{Planning Agent} identifies the user's complex query and breaks it down into manageable sub-tasks. \textbf{Executor Agents} are then tasked with collecting task specific information, utilizing access to external tools. \textbf{Answer Agent} compiles the findings into a cohesive response, which is subsequently verified by the \textbf{Verifier Agent} to ensure accuracy and relevance. 

By dividing tasks among different agents, each agent can focus on a specific part of the task, accessing only the necessary history as context during task execution, which effectively addresses the issue of \textbf{redundant context}. The redesign of the rollback mechanism to incorporate memory and tool list rollback mechanisms addresses the \textbf{instability of the rollback mechanism}. Drawing on the concept of least-to-most prompting~\cite{zhou2023leasttomost}, the original problem is first decomposed into sub-problems for macro-level planning. Subsequently, Smurfs is used to solve each sub-problem at the micro-level, with macro-level planning guiding the micro-level planning, thereby resolving the issue of \textbf{premature termination}. 

In the rest of this section, the mechanism of the system and the functions of each agent will be detailed. More details of memory system can be seen at Appendix~\ref{appendix: details smurfs}.

\subsection{Agent Components}
In this section, we introduce the two core components of the Smurfs system:

\paragraph{Tools}

The tool documents about the tools that Smurfs can utilize in the completion of a complex task are denoted as $D=\{n_{i}, d_{i}, p_{i}\}^{|d|}_{i=1}$, where n represents the tool name, d represents tool usage description, p represents parameter description and $|d|$ represents the amount of the available tools. The available tool list is denoted as $\tau = \{n_{i}, d_{i}\}^{|\tau|}_{i=1}$. $\tau_{t}$ denotes the tool list Smurfs can utilize at time t.

\paragraph{Memory}
The memory of the agent system at time t is the history of the task-solving process before t, denoted as $M=(m_{1}, m_{2}, ..., m_{t-1})$ and $m_{i}=(\gamma_{i}, a_{i})$, where $m_{i}$ represents memory element at time i and $\gamma_{i}$, $a_{i}$ represents thought and answer generated by the system at time i. There are two types of memory in Smurfs: local memory and global memory. the local memory is used to record the ongoing solution trajectory and to generate the next action in the current trajectory. The global memory, meanwhile, records all trajectories and is used to generate the sub-problem's answer by combining all trajectory records when the maximum number of retries is exceeded. This local-global combined memory system ensures that the planning of the current solution trajectory is not influenced by the context of erroneous trajectories. It also generates an answer that combines all trajectories when the verifier agent cannot determine task completion within the maximum number of planned steps. This memory system ensures context efficiency during the task-solving process.

\subsection{Macro Planning}
\paragraph{Planning Agent}

The primary responsibility of the Planning Agent is doing macro-level planning through task decomposition to prevent \textbf{premature termination}. The inference process of the Planning Agent is:
\begin{equation}
    Plan \: P:(p1, p2, ...)=PA(q)    
\end{equation}
Where $p_{i}$ represents sub-problem of the original query q, PA represents the Planning Agent. After the task decomposition, the agent system will use Executor Agent, Answer Agent an Verifier Agent to solve each sub-problem using DFSDT collaboratively in a sequential order. To utilize the answer of the previous sub-problem when solving subsequent sub-problem, the strategy known as least-to-most prompting~\cite{zhou2023leasttomost} is used. 

\subsection{Subtask Solving Process}

After introducing the function of plan agent, this section outlines how the agents collaborate to solve sub-tasks, as shown in Figure~\ref{fig:multi-agent_detail1}.
\paragraph{Stable Rollback}

To address the \textbf{instability of the rollback mechanism} in DFSDT, we propose a rollback mechanism based on rules. Whenever an error occurs while using a tool $\tau_{t, i}$ at time t, the tool list at t $\tau_{t}$ will pop $\tau_{t, i}$ out and reperform tool selection and tool planning (ensuring that the faulty tool is not selected again). If, at time t, the tool list becomes empty, it signifies that after the system choosing tool $\tau_{t-1, j}$ at time t-1, no subsequent trajectory can solve the problem. In this case, the agent system will roll back to time t-1, meaning that the local memory M will pop out the memory element $m_{t-1}$ at time t-1, and the tool list at time t-1 $\tau_{t-1}$ will pop out tool $\tau_{t-1, j}$. The agent system will then set the time t=t-1 and continue planning. This rule-based rollback mechanism, compared to the original model-based rollback mechanism of DFSDT, is less flexible and might reduce rollback efficiency. However, it is more stable, ensuring the correctness of deep first search and enabling models with weaker capabilities to utilize DFSDT on tool planning.

\begin{table*}[htp]
\centering
\footnotesize
\scalebox{0.70}{
    \begin{tabular}{lc|cc|cc|cc|cc|cc|cc|cc}
    \toprule
    \multirow{3}{*}{\textbf{Backbone}}  &  \multirow{3}{*}{\textbf{Method}} & \multicolumn{14}{c}{\textbf{StableToolBench}} \\
    & & \multicolumn{2}{c}{\textbf{I1-Inst.}} & \multicolumn{2}{c}{\textbf{I1-Cat.}} & \multicolumn{2}{c}{\textbf{I1-Tool.}} & \multicolumn{2}{c}{\textbf{I2-Cat.}}   & \multicolumn{2}{c}{\textbf{I2-Inst.}} & \multicolumn{2}{c}{\textbf{I3-Inst.}} & \multicolumn{2}{c}{\textbf{Average}} \\
    & & Pass & Win & Pass & Win & Pass & Win & Pass & Win & Pass & Win & Pass & Win & Pass & Win \\
    \midrule
           GPT-3.5 Turbo & ReACT & $41.6_{\pm 1.2}$ & / & $48.4_{\pm 0.5}$ & / & $52.5_{\pm 0.5}$ & / & $52.2_{\pm 1.0}$ & / & $31.6_{\pm 1.2}$ & / & $39.9_{\pm 2.0}$ & / & $44.4_{\pm 1.1}$ & $/$ \\
           GPT-3.5 Turbo & DFSDT & $54.1_{\pm 1.0}$ & $64.4$ & $60.1_{\pm 0.0}$ & $61.4$ & $59.9_{\pm 1.7}$ & $53.8$ & $60.9_{\pm 0.9}$ & $62.9$ & $52.8_{\pm 3.7}$ & $66.0$ & $44.3_{\pm 4.8}$ & $54.1$ & $55.4_{\pm 2.0}$ & $60.4$ \\
           \rowcolor{c5!50} GPT-3.5 Turbo & Smurfs & $\underline{60.3_{\pm 1.5}}$ & $65.0$ & $67.0_{\pm 1.0}$ & $\underline{69.9}$ & $60.3_{\pm 1.3}$ & $54.4$ & $54.3_{\pm 0.4}$ & $\underline{63.7}$ & $42.6_{\pm 1.6}$ & $64.2$ & $60.1_{\pm 1.0}$ & $57.4$ & $57.4_{\pm 1.1}$ & $62.4$ \\
           Mistral-7B & ReACT & 0.0 & 0.0 & 0.0 & 0.0 & 0.0 & 0.0 & 0.0 & 0.0 & 0.0 & 0.0 & 0.0 & 0.0 & 0.0 & 0.0 \\
           Mistral-7B & DFSDT & 0.0 & 0.0 & 0.0 & 0.0 & 0.0 & 0.0 & 0.0 & 0.0 & 0.0 & 0.0 & 0.0 & 0.0 & 0.0 & 0.0 \\
           \rowcolor{c5!50} Mistral-7B & Smurfs & $\mathbf{76.3_{\pm 0.8}}$ & 63.8 & $\mathbf{86.7_{\pm 1.2}}$ & 62.7 & $\mathbf{81.0_{\pm 1.9}}$ & 58.2 & $\mathbf{70.4_{\pm 2.7}}$ & 54.0 & $\mathbf{63.8_{\pm 2.4}}$ & $\underline{67.0}$ & $\mathbf{85.2_{\pm 0.7}}$ & $57.4$ & $\mathbf{77.2_{\pm 1.6}}$ & $60.5$ \\
           GPT-4 Turbo & ReACT & $41.1_{\pm 1.5}$ & $60.1$ & $53.2_{\pm 1.3}$ & $62.1$ & $42.2_{\pm 1.1}$ & $48.1$ & $50.0_{\pm 0.7}$ & $57.3$ & $38.7_{\pm 0.8}$ & $65.1$ & $37.7_{\pm 1.3}$ & $47.5$ & $43.8_{\pm 1.1}$ & $56.7$ \\
           GPT-4 Turbo & DFSDT & $52.7_{\pm 1.4}$ & $\underline{69.9}$ & $58.2_{\pm 0.9}$ & $66.0$ & $59.7_{\pm 1.2}$ & $\underline{58.2}$ & $59.3_{\pm 0.7}$ & $62.1$ & $52.2_{\pm 2.3}$ & $\mathbf{67.9}$ & $61.5_{\pm 1.8}$ & $\underline{65.6}$ & $57.3_{\pm 1.4}$ & $\underline{65.0}$ \\
           \rowcolor{c5!50} GPT-4 Turbo & Smurfs & $59.3_{\pm 1.4}$ & $\mathbf{71.2}$ & $\underline{73.3_{\pm 1.3}}$ & $\mathbf{72.5}$ & $\underline{67.4_{\pm 0.7}}$ & $\mathbf{69.6}$ & $\underline{66.7_{\pm 1.9}}$ & $\mathbf{73.4}$ & $\underline{55.5_{\pm 1.4}}$ & $66.0$ & $\underline{70.5_{\pm 0.0}}$ & $\mathbf{72.1}$ & $\underline{65.5_{\pm 1.1}}$ & $\mathbf{70.8}$ \\
    \bottomrule
    \end{tabular}
}
\caption{The open-end tool planning task evaluation on the StableToolBench benchmark~\cite{guo2024stabletoolbench}. The most effective approach is highlighted in bold, while the second-best is underlined. Win rate is calculated by comparing each model with ChatGPT-ReACT. A win rate higher than 50\% means the model performs better than ChatGPT-ReACT.}
\label{tab:Performance Comparison of LLMs on ToolBench}
\end{table*}

\paragraph{Executor Agent}


The Executor Agent is responsible for choosing and executing the tools to solve the sub-tasks. At each time t, the agent can invoke one tool to tackle the given task:  
\begin{align}
    \gamma=EA.gen\_thought(p, M, \tau, h) \\
    \alpha=EA.choose\_tool(p, \gamma, \tau) \\
    \beta=EA.gen\_arguments(p, M, D[\alpha]) \\
    r=EA.call\_tool(\alpha, \beta)
\end{align}
Where p is the sub-problem from Planning Agent, h is the hint from the Verifier Agent, $\tau$ is the tool list, M is local memory, $D[\alpha]$ means the tool document of tool $\alpha$. The agent, using the ReACT format~\cite{yao2022react} to choose the tool and arguments, then execute the tool. Noticed that each inference process only uses the relevant part from the local memory and tool list to reduce the context redundancy. More detailed information of the Executor Agent can be found in Figure~\ref{fig:multi-agent_detail3}.

\paragraph{Answer Agent}

To mitigate the performance degradation caused by lengthy contexts, we introduce the Answer Agent role, designed to extract crucial content for each step and sub-problem:
\begin{align}
    Answer: a = AA(q, r, M)
\end{align}
Where q is sub-problem from the Planning Agent, r is response from the Executor Agent, M is the local memory (or global memory if max retry reaches). 
As the `lost-in-the-middle' theory described in section~\ref{sec: intro},
retaining all information may not always be beneficial, particularly in cases where the solution path is challenging to discern. Therefore, the primary role of the Answer Agent is to succinctly summarize the generated answers and tool responses to maintain the memory efficiency.

\paragraph{Verifier Agent}

The Verifier Agent serves as an early-stopping and reflection mechanism, allowing for a balance between effectiveness and efficiency:
\begin{align}
    h, c = VA(q, a)
\end{align}
Where q denotes the sub-problems from the Planning Agent, a denotes the answer from the answer agent, h and c denotes hint and check status respectively. If check status generated is 0, that means the Verifier Agent thinks the sub-problem isn't completed, the system will add the thought and answer of this time to the local and global memory, set t=t+1 and continue the inference procedure.If check status is 1, the sub-problems is thought to be solved and the system will start to deal with the next sub-problem.

\section{Experiments}
\label{sec: experiments}



To evaluate both the effectiveness and efficiency of the Smurfs framework, in thie section, we carried out two multi-tool planning tasks: (1) an open-ended task, \textit{StableToolBench}~\cite{guo2024stabletoolbench}, and (2) a closed-ended task, \textit{HotpotQA}~\cite{yang2018hotpotqa}. In addition to these main experiments designed to assess the entire framework, we conducted an ablation studies followed by a case study to test the capabilities of each component within the multi-agent framework and investigate the underlying reasons for its effectiveness.






\subsection{Open-ended Task: StableToolBench}

StableToolBench is a tool learning benchmark derived from ToolBench~\cite{qin2024toolllm}, encompassing multi-step tool usage tasks across over 16,000 APIs. The benchmark employs two metrics for evaluation:
\textbf{(1) Pass Rate} measures the percentage of instructions successfully executed within the allocated budget.
\textbf{(2) Win Rate} represents the preference selection by a ChatGPT evaluator when presented with two solution paths.

\paragraph{Baselines} Following the original paper that introduced the benchmark, we adopt \textit{ReACT}~(\textit{CoT})~\cite{wei2023chainofthought} and \textit{DFSDT}~\cite{touvron2023llama} as baseline methods for comparison. Additionally, we include the backbones used in the paper: gpt-3.5-turbo-0613 (GPT-3.5 Turbo)~\cite{ChatGPT} and gpt-4-turbo-preview (GPT-4 Turbo). To explore the adaptability of the tool-planning methods, we also include Mistral-7B-Instruct-v0.2 (Mistral-7B)~\cite{jiang2023mistral} as one of the selected backbones in our experiments.

\paragraph{Settings} To minimize the influence of varying tool APIs on experimental results, we conducted all experiments using the same API cache~\cite{guo2024stabletoolbench}. For a fair comparison among the candidate methods and to reduce variability, each model was executed once and evaluated three times, with results averaged. Other settings follow those specified in the original benchmark paper.

\paragraph{Results} Table~\ref{tab:Performance Comparison of LLMs on ToolBench} displays the results on StableToolBench. For the untrained LLM, Mistral-7B, existing agent frameworks did not improve its performance in tool planning tasks; Mistral-7B failed these tasks when integrated with the ReACT and DFSDT frameworks~\footnote{Experiment results show that Mistral-7B failed to correctly execute the `finish' action during inference, resulting in invalid responses.}. However, Smurfs exhibited exceptional performance: when combined with Mistral-7B, Smurfs achieved competitive scores among the baselines. Through its task decomposition mechanism, Smurfs transforms long-context tasks into simpler ones, enabling the untrained model to effectively utilize external tools for managing complex tasks. 
Regarding closed-source models, specifically GPT4 in these experiments, Smurfs also demonstrated outstanding performance on the benchmark compared to other agent frameworks and achieved state-of-the-art results on the benchmark. Its high success rate suggests that Smurfs is more effective at finding optimal solution paths compared to ChatGPT.

\paragraph{Further Analysis} 
We conducted a detailed analysis of the token costs associated with each tool planning method for the tasks, a critical evaluation aspect for multihop reasoning tasks. As shown in Table~\ref{tab:token cost} (detailed in Appendix~\ref{appendix: token cost}), the average token costs per question and API request are evaluated for ReACT, DFSDT, and Smurfs on StableToolBench.
The analysis reveals that DFSDT generally requires about 20,000 tokens per question, encompassing both prompt and completion tokens. This is nearly three times the token cost compared to ReACT and twice as much as Smurfs. Despite this higher token cost, DFSDT does not demonstrate commensurate effectiveness improvements over other methods. 
These findings underscore the cost-efficiency of the proposed MAS framework, Smurfs, which not only reduces token expenditure in solving multihop planning tasks but also delivers outstanding performance in evaluations.

\begin{table}[ht]
\centering
\footnotesize
\scalebox{0.75}{
    \begin{tabular}{cl|cccc}
    \toprule
    \multirow{2}{*}{\textbf{Backbone}}  &  \multirow{2}{*}{\textbf{Method} \, \makecell[c]{{\tiny \faUser}{\tiny Single-Agent} \\ {\tiny \faUsers} {\tiny Multi-Agent}} }  & \multicolumn{4}{c}{\textbf{HotpotQA}} \\
            &           & \textbf{Easy} & \textbf{Medium} & \textbf{Hard} & \textbf{All} \\
    \midrule
    \multirow{2}{*}{\makecell{GPT-3.5\\Turbo}}   
            & {\tiny \faToggleOff} {\tiny \faUser} CoT & 48.21 & 44.52 & 34.22 & 42.32 \\
            & {\tiny \faToggleOff} {\tiny \faUser} Zero-Shot Plan & 50.71 & 45.17 & 38.23 & 44.70 \\
    \midrule
    \multirow{8}{*}{\makecell{Mistral-7B\\Instruct-v0.2}}   
            & {\tiny \faToggleOff} {\tiny \faUser} CoT & 33.70 & 22.38 & 22.14 & 26.07 \\
            & {\tiny \faToggleOff} {\tiny \faUser} ReAct & 38.09 & 27.57 & 22.05 &  29.24 \\
            & {\tiny \faToggleOff} {\tiny \faUser} Chameleon & 37.07 &  26.67 & 19.20 & 27.65 \\
            & {\tiny \faToggleOff} {\tiny \faUser} Reflexion & 40.78 & 35.02 & 28.36 & 34.72 \\
            & {\tiny \faToggleOff} {\tiny \faUsers} BOLAA & 40.86 & 32.11 & 22.36 &  31.78 \\
            & {\tiny \faToggleOff} {\tiny \faUsers} ReWOO & 38.42 &  31.89 & 25.98 & 32.10 \\
            & \cellcolor{c5!50}{\tiny \faToggleOff} {\tiny \faUsers} Smurfs (ours) & \cellcolor{c5!50}\underline{45.94} & \cellcolor{c5!50}\textbf{40.74} & \cellcolor{c5!50}\underline{30.72} & \cellcolor{c5!50}\textbf{39.13} \\
            & {\tiny \faToggleOn} {\tiny \faUser} FireAct & 45.52 & 32.02 & 30.17 & 35.90 \\
            & {\tiny \faToggleOn} {\tiny \faUsers} AUTOACT & \textbf{48.69} & \underline{36.65} & \textbf{31.37} & \underline{38.89} \\
    \midrule
    \multirow{8}{*}{\makecell{Llama-2\\13B-chat}}
            & {\tiny \faToggleOff} {\tiny \faUser} CoT & 37.90 & 25.28 & 21.64 & 28.27 \\
            & {\tiny \faToggleOff} {\tiny \faUser} ReAct &  28.68 &  22.15 &  21.69 &  24.17 \\
            & {\tiny \faToggleOff} {\tiny \faUser} Chameleon & 40.01 & 25.39 & 22.82 & 29.41 \\
            & {\tiny \faToggleOff} {\tiny \faUser} Reflexion &  44.43 &  37.50 &  \underline{28.17} & 36.70 \\
            & {\tiny \faToggleOff} {\tiny \faUsers} BOLAA & 33.23 & 25.46 & 25.23 &  27.97 \\
            & {\tiny \faToggleOff} {\tiny \faUsers} ReWOO &  30.09 & 24.01 & 21.13 & 25.08 \\
            & \cellcolor{c5!50}{\tiny \faToggleOff} {\tiny \faUsers} Smurfs (ours) & \cellcolor{c5!50}42.62 & \cellcolor{c5!50}27.21 & \cellcolor{c5!50}22.92 & \cellcolor{c5!50}30.92 \\
            & {\tiny \faToggleOn} {\tiny \faUser} FireAct &  \underline{45.83} & \underline{38.94} & 26.06 & \underline{36.94} \\
            & {\tiny \faToggleOn} {\tiny \faUsers} AUTOACT & \textbf{47.29} & \textbf{41.27} & \textbf{32.92} & \textbf{40.49} \\
    \midrule
    \multirow{8}{*}{\makecell{Llama-2\\70B-chat}}   
            & {\tiny \faToggleOff} {\tiny \faUser} CoT &  45.37 &  36.33 &  32.27 & 37.99 \\
            & {\tiny \faToggleOff} {\tiny \faUser} ReAct &  39.70 &   37.19 &  33.62 &  36.83 \\
            & {\tiny \faToggleOff} {\tiny \faUser} Chameleon & 46.86 & 38.79 & 34.43 & 40.03 \\
            & {\tiny \faToggleOff} {\tiny \faUser} Reflexion & 48.01 & 46.35 & 35.64 & 43.33 \\
            & {\tiny \faToggleOff} {\tiny \faUsers} BOLAA & 46.44 & 37.29 & 33.49 & 39.07 \\
            & {\tiny \faToggleOff} {\tiny \faUsers} ReWOO & 42.00 & 39.58 & 35.32 & 38.96 \\
            & \cellcolor{c5!50}{\tiny \faToggleOff} {\tiny \faUsers} Smurfs (ours) & \cellcolor{c5!50}\underline{52.86} & \cellcolor{c5!50}\textbf{50.77} & \cellcolor{c5!50}\textbf{44.87} & \cellcolor{c5!50}\textbf{49.50} \\
            & {\tiny \faToggleOn} {\tiny \faUser} FireAct & 50.82 & 41.43 & 35.86 & 42.70 \\
            & {\tiny \faToggleOn} {\tiny \faUsers} AUTOACT & \textbf{56.94} & \underline{50.12} & \underline{38.35} & \underline{48.47} \\
    \bottomrule
    \end{tabular}
}
\caption{The closed-end tool planning evaluation on HotpotQA~\cite{yang2018hotpotqa}, with some results derived from \cite{qiao2024autoact}. The most effective approach for each group is highlighted in bold, while the second-best is underlined. Methods marked with \faToggleOn require model training.}
\label{tab:Performance Comparison of LLMs on HotpotQA}
\end{table}

\subsection{Closed-ended Task: HotpotQA}
Compared to open-ended tasks, closed-ended tasks provide a more stable and robust evaluation. To this end, we evaluate the methods on HotpotQA~\cite{yang2018hotpotqa} in addition to StableToolBench. HotpotQA is a multi-hop QA task that is challenging due to the requirement for rich background knowledge, with answers typically being short entities or yes/no responses. 

\paragraph{Baselines}
The compared baselines include \textbf{CoT} \cite{wei2023chainofthought}, \textbf{\textsc{ReAct}}\cite{yao2022react}, \textbf{Chameleon}\cite{lu2023chameleon}, \textbf{Reflexion} \cite{shinn2023reflexion}, \textbf{BOLAA} \cite{liu2023bolaa}, \textbf{ReWOO} \cite{xu2023rewoo}, \textbf{\textsc{FireAct}} \cite{chen2023fireact}, \textbf{AutoAct}\cite{qiao2024autoact}.

\paragraph{Settings and Metrics}
Following the settings in ~\cite{qiao2024autoact}, we use open-source Llama-2 models~\cite{touvron2023llama} and Mistral-7B~\cite{jiang2023mistral} as the backbones of each agent to evaluate the performance of Smurfs. The evaluation metrics is $\texttt{reward} \in [0,1]$, defined as the F1 score grading between the prediction and ground-truth answer. For more details about the experiment, see Appendix~\ref{appendix: experiment settings}.


\paragraph{Results}
Smurfs, as an untrained MAS system, not only comprehensively outperforms untrained agents but also achieves and even surpasses the accuracy of trained agents across most backbone models. This sufficiently demonstrates that the mechanism of smurfs ensures strong generalization capabilities while maintaining high effectiveness. 

Observations indicate that the performance of LLama-2-13b-chat on smurfs-related tasks is suboptimal, likely due to its limited capabilities in tool arguments generation. Specifically, the primary issue identified is that, when the Executor agent successfully selects relevant tool, it tends to produce hallucination arguments that can't be used by the tools. This indicates that LLama-2-13b-chat may need further training for usage of tools. The experiment results may substantiate this viewpoint, demonstrating that the untrained methods of llama-2-13b-chat generally exhibit significantly lower accuracy compared to the trained methods. To provide further insight into the model's performance in the tool planning process, we manually categorized the types of errors made by Smurfs on the hotpotQA hard dataset in appendix~\ref{appendix: Error}.

\begin{table}[h]
    \centering
    \footnotesize
    \scalebox{1.0}{
        \begin{tabular}{l|cc}
        \toprule
            & \multicolumn{2}{c}{I3-Inst.} \\
            & Pass ($\%$) & Win ($\%$) \\
        \midrule
        GPT-3.5 Turbo \textit{with Smurfs} & $60.1_{\pm 1.0}$ & 57.4 \\
        \textit{\quad w/o Answer Agent} & $57.4_{\pm 2.9}$ & $49.2$ \\
        \textit{\quad w/o Verifier Agent} & $54.1_{\pm 2.7}$ & $42.6$ \\
        \textit{\quad w/o Planning Agent} & $35.5_{\pm 3.3}$ & $42.6$ \\
        \textit{\quad w/o Planning \& Verifier Agent} & $58.5_{\pm 2.0}$ & $57.4$ \\
        \midrule
        GPT-4 Turbo \textit{with Smurfs} & $70.5_{\pm 1.0}$ & 72.1 \\
        \textit{\quad w/o Answer Agent} & $82.2_{\pm 2.5}$ & $72.1$ \\
        \textit{\quad w/o Verifier Agent} & $79.2_{\pm 0.8}$ & $63.9$ \\
        \textit{\quad w/o Planning Agent} & $71.9_{\pm 2.8}$ & $63.9$ \\
        \textit{\quad w/o Planning \& Verifier Agent} & $79.8_{\pm 2.8}$ & $67.2$ \\
        \bottomrule
        \end{tabular}
    }
    \caption{Ablation study on StableToolBench I3-Inst subset to investigate the importance of each component within the framework.}
    \label{tab:ablation study}
\end{table}


\begin{figure*}[htp]
    \centering
    \includegraphics[width=0.8\textwidth]{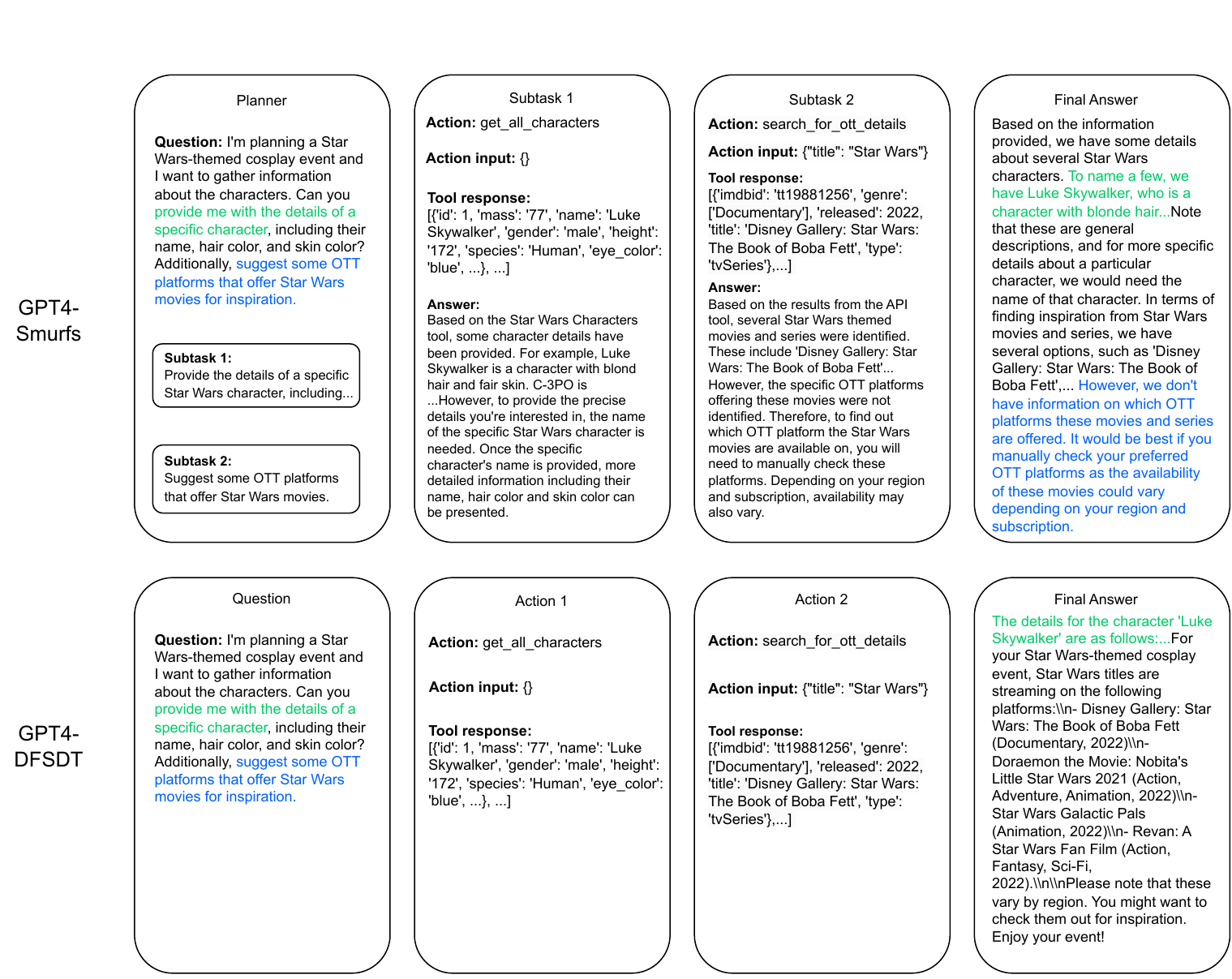}
    \caption{The illustration of how GPT4-Smurfs and GPT4-DFSDT solve long context problem. The two sub-questions and their corresponding answers are marked in two colors.}
    \label{fig:case-study-1}
\end{figure*}

\section{Ablation Study}

\subsection{Importance of each component in MAS}


We performed an ablation study to investigate the impact of each agent in our framework. We removed each agent individually, except for the indispensable Executor Agent, and compared the results to the complete framework.

\paragraph{Settings}
\textbf{(1) Verifier Agent Removal:} Without Verifier Agent, the framework uses a generatl DFSDT in each subtask, i.e. including a finish tool into the tool library and deciding whether to stop at tool choice step.
\textbf{(2) Answer Agent Removal:} Without Answer Agent, the framework uses full tool response instead of the summary of tool response in its memory.
\textbf{(3) Planning Agent Removal:} Without Planning Agent, the framework uses Verifier Agent to decide whether the task is completed.
\textbf{(4) Planning \& Verifier Agent Removal:} Without Planning and Verifier Agent, the only difference between Smurfs and DFSDT will be the Answer Agent, i.e. including a finish tool in the tool library, deciding whether to stop at tool choice step and using summary of tool response in its memory.
\paragraph{Results}
Table~\ref{tab:ablation study} shows the experiment result, highlighting several key insights regarding the impact of different components in the multi-agent system.

First, the removal of any component generally reduces the win rate, highlighting each component's significance. Second, performance degradation trends remain consistent: Eliminating the Answer Agent has minimal impact, whereas removing the Planning Agent causes the greatest decline. However, this does not imply the relative importance of these agents but rather suggests that the Verifier-Planning Agent workflow is more robust than the Verifier-Answer Agent workflow.

Third, the performance impact varies by model capability. For GPT-3.5, the removal of the Answer Agent reduces both the pass and the win rates, while for GPT-4, it preserves the win rate and improves the pass rate, probably due to the superior long-text reasoning ability of GPT-4, which processes additional tool response more effectively. Removal of the Planning Agent significantly reduces the pass rate of GPT-3.5 but has a minimal effect on GPT-4, suggesting that the GPT-4 Verifier Agent is more robust. \textbf{These findings may indicate that more powerful models can compensate for agent removal, sometimes even benefiting from additional context, while weaker models benefit more from complex workflow and compressed context.}


We also noticed an interesting phenomenon: when only the Planning Agent is removed, the system experiences a major decline in performance, while removing the Verifier Agent in addition to the Planning Agent improves performance again. This may indicate that the Verifier Agent’s ability to pause tasks or provide next-step guidance is suboptimal and requires further training.

In conclusion, our findings offer some insight into the relationship between model capabilities and multi-agent system performance. The optimal number of agents and workflow may vary depending on the capacity of the model. \textbf{We propose the hypothesis that weaker models benefit more from complex multi-agent systems and context segmentation, while stronger models perform better with comprehensive context and simpler agent systems.} To validate this hypothesis, more extensive ablation studies are needed under a wider range of models and constraints to explore the influence of different context structures and reasoning workflows. We leave this investigation for future work.




\subsection{Case Study}
As shown in Figure~\ref{fig:case-study-1}, although GPT4-DFSDT and GPT4-Smurfs use the same tool calls to solve the problem, GPT4-DFSDT only answers the first sub-question correctly while GPT4-Smurfs answers both sub-questions accurately. In the process of addressing the second sub-question, it is notable that the tool response only mentions titles of film and television products related to "Star Wars", without addressing OTT platforms. GPT-4-DFSDT erroneously interprets these titles as responses to the question, while GPT-4-Smurfs adeptly identifies this discrepancy and provides a more appropriate response. This case highlights that in situations where tool responses are lengthy and questions are complex, the single agent framework like DFSDT may be susceptible to distractions from irrelevant information, leading to erroneous answers. Conversely, the context-efficient Smurfs framework demonstrates a reduced susceptibility to irrelevant information, thereby generating more accurate answers.

\section{Conclusion}
\label{sec: conclusion}


In this paper, we propose \textbf{\textit{Smurfs}}, an innovative MAS framework designed to enhance the tool-planning capabilities of LLMs without requiring additional training. Through extensive experiments on both open-ended and closed-ended tool planning benchmarks, Smurfs demonstrate its effectiveness by consistently outperforming baseline methods. Ablation study further provides deeper insights into the impact of each agent in our framework. Based on the ablation study, we propose a hypothesis that can be further tested through more comprehensive experiments in future research: weaker models benefit more from complex multi-agent systems and context segmentation, while stronger models perform better with comprehensive context and simpler agent systems. The findings not only advance the state-of-the-art in multi-tool planning systems but also highlight the potential of modular, training-free frameworks for LLMs in various practical applications.

Looking forward, future research could focus on exploring Smurfs' use in new domains, such as facilitating the synthesis of high-quality multi-tool planning data and enhancing the base model's reasoning and tool-use abilities, further advancing the field of adaptive AI systems.



\section*{Limitations}

\paragraph{\textbf{Model Size Constraints: }} Due to computational constraints, our experiments did not include larger and more diverse types of LLMs. We believe this would not affect the main observations of this paper.

\paragraph{\textbf{Agent Component Scale-Up: }} Although we selected the most common and intuitive agent roles for the proposed MAS, there are many possibilities for researchers to explore. Investigating more well-designed agent roles may help improve the effectiveness of the agent system, and developing automated methods to identify these roles could facilitate effective scaling.

Acknowledging these limitations, future research should aim to address these gaps to provide a more comprehensive understanding of the Smurfs framework's capabilities and potential areas for improvement.

\section*{Acknowledgements}
This work was supported by  the Shenzhen Science and Technology Program (JCYJ20220818103001002), Shenzhen Doctoral Startup Funding (RCBS20221008093330065), Tianyuan Fund for Mathematics of National Natural Science Foundation of China (NSFC) (12326608), Shenzhen Key Laboratory of Cross-Modal Cognitive Computing (grant number ZDSYS20230626091302006), and Shenzhen Stability Science Program 2023.
\bibliography{custom}

\begin{thebibliography}{36}
\providecommand{\natexlab}[1]{#1}

\bibitem[{Ambrose(2001)}]{ambrose2001paleolithic}
Stanley~H Ambrose. 2001.
\newblock Paleolithic technology and human evolution.
\newblock \emph{Science}, 291(5509):1748--1753.

\bibitem[{Chen et~al.(2023{\natexlab{a}})Chen, Shu, Shareghi, Collier, Narasimhan, and Yao}]{chen2023fireact}
Baian Chen, Chang Shu, Ehsan Shareghi, Nigel Collier, Karthik Narasimhan, and Shunyu Yao. 2023{\natexlab{a}}.
\newblock \href {https://arxiv.org/abs/2310.05915} {Fireact: Toward language agent fine-tuning}.
\newblock \emph{Preprint}, arXiv:2310.05915.

\bibitem[{Chen et~al.(2023{\natexlab{b}})Chen, Su, Zuo, Yang, Yuan, Chan, Yu, Lu, Hung, Qian, Qin, Cong, Xie, Liu, Sun, and Zhou}]{chen2023agentverse}
Weize Chen, Yusheng Su, Jingwei Zuo, Cheng Yang, Chenfei Yuan, Chi-Min Chan, Heyang Yu, Yaxi Lu, Yi-Hsin Hung, Chen Qian, Yujia Qin, Xin Cong, Ruobing Xie, Zhiyuan Liu, Maosong Sun, and Jie Zhou. 2023{\natexlab{b}}.
\newblock \href {https://arxiv.org/abs/2308.10848} {Agentverse: Facilitating multi-agent collaboration and exploring emergent behaviors}.
\newblock \emph{Preprint}, arXiv:2308.10848.

\bibitem[{Dai et~al.(2019)Dai, Yang, Yang, Carbonell, Le, and Salakhutdinov}]{dai2019transformer}
Zihang Dai, Zhilin Yang, Yiming Yang, Jaime Carbonell, Quoc~V Le, and Ruslan Salakhutdinov. 2019.
\newblock Transformer-xl: Attentive language models beyond a fixed-length context.
\newblock \emph{arXiv preprint arXiv:1901.02860}.

\bibitem[{Dao et~al.(2022)Dao, Fu, Ermon, Rudra, and R{\'e}}]{dao2022flashattention}
Tri Dao, Dan Fu, Stefano Ermon, Atri Rudra, and Christopher R{\'e}. 2022.
\newblock Flashattention: Fast and memory-efficient exact attention with io-awareness.
\newblock \emph{Advances in Neural Information Processing Systems}, 35:16344--16359.

\bibitem[{Fu et~al.(2024)Fu, Anantha, and Cheng}]{fu2024camphorcollaborativeagentsmultiinput}
Yicheng Fu, Raviteja Anantha, and Jianpeng Cheng. 2024.
\newblock \href {https://arxiv.org/abs/2410.09407} {Camphor: Collaborative agents for multi-input planning and high-order reasoning on device}.
\newblock \emph{Preprint}, arXiv:2410.09407.

\bibitem[{Guo et~al.(2024)Guo, Cheng, Wang, Liang, Qin, Li, Liu, Sun, and Liu}]{guo2024stabletoolbench}
Zhicheng Guo, Sijie Cheng, Hao Wang, Shihao Liang, Yujia Qin, Peng Li, Zhiyuan Liu, Maosong Sun, and Yang Liu. 2024.
\newblock \href {https://arxiv.org/abs/2403.07714} {Stabletoolbench: Towards stable large-scale benchmarking on tool learning of large language models}.
\newblock \emph{Preprint}, arXiv:2403.07714.

\bibitem[{Hong et~al.(2023)Hong, Zheng, Chen, Cheng, Wang, Zhang, Wang, Yau, Lin, Zhou et~al.}]{hong2023metagpt}
Sirui Hong, Xiawu Zheng, Jonathan Chen, Yuheng Cheng, Jinlin Wang, Ceyao Zhang, Zili Wang, Steven Ka~Shing Yau, Zijuan Lin, Liyang Zhou, et~al. 2023.
\newblock Metagpt: Meta programming for multi-agent collaborative framework.
\newblock \emph{arXiv preprint arXiv:2308.00352}.

\bibitem[{Jiang et~al.(2023)Jiang, Sablayrolles, Mensch, Bamford, Chaplot, Casas, Bressand, Lengyel, Lample, Saulnier et~al.}]{jiang2023mistral}
Albert~Q Jiang, Alexandre Sablayrolles, Arthur Mensch, Chris Bamford, Devendra~Singh Chaplot, Diego de~las Casas, Florian Bressand, Gianna Lengyel, Guillaume Lample, Lucile Saulnier, et~al. 2023.
\newblock Mistral 7b.
\newblock \emph{arXiv preprint arXiv:2310.06825}.

\bibitem[{Li et~al.(2023)Li, Hammoud, Itani, Khizbullin, and Ghanem}]{li2023camelcommunicativeagentsmind}
Guohao Li, Hasan Abed Al~Kader Hammoud, Hani Itani, Dmitrii Khizbullin, and Bernard Ghanem. 2023.
\newblock \href {https://arxiv.org/abs/2303.17760} {Camel: Communicative agents for "mind" exploration of large language model society}.
\newblock \emph{Preprint}, arXiv:2303.17760.

\bibitem[{Liu et~al.(2024)Liu, Lin, Hewitt, Paranjape, Bevilacqua, Petroni, and Liang}]{liu2024lost}
Nelson~F Liu, Kevin Lin, John Hewitt, Ashwin Paranjape, Michele Bevilacqua, Fabio Petroni, and Percy Liang. 2024.
\newblock Lost in the middle: How language models use long contexts.
\newblock \emph{Transactions of the Association for Computational Linguistics}, 12:157--173.

\bibitem[{Liu et~al.(2023)Liu, Yao, Zhang, Xue, Heinecke, Murthy, Feng, Chen, Niebles, Arpit, Xu, Mui, Wang, Xiong, and Savarese}]{liu2023bolaa}
Zhiwei Liu, Weiran Yao, Jianguo Zhang, Le~Xue, Shelby Heinecke, Rithesh Murthy, Yihao Feng, Zeyuan Chen, Juan~Carlos Niebles, Devansh Arpit, Ran Xu, Phil Mui, Huan Wang, Caiming Xiong, and Silvio Savarese. 2023.
\newblock \href {https://arxiv.org/abs/2308.05960} {Bolaa: Benchmarking and orchestrating llm-augmented autonomous agents}.
\newblock \emph{Preprint}, arXiv:2308.05960.

\bibitem[{Lu et~al.(2023)Lu, Peng, Cheng, Galley, Chang, Wu, Zhu, and Gao}]{lu2023chameleon}
Pan Lu, Baolin Peng, Hao Cheng, Michel Galley, Kai-Wei Chang, Ying~Nian Wu, Song-Chun Zhu, and Jianfeng Gao. 2023.
\newblock \href {https://arxiv.org/abs/2304.09842} {Chameleon: Plug-and-play compositional reasoning with large language models}.
\newblock \emph{Preprint}, arXiv:2304.09842.

\bibitem[{Mallen et~al.(2022)Mallen, Asai, Zhong, Das, Khashabi, and Hajishirzi}]{mallen2022not}
Alex Mallen, Akari Asai, Victor Zhong, Rajarshi Das, Daniel Khashabi, and Hannaneh Hajishirzi. 2022.
\newblock When not to trust language models: Investigating effectiveness of parametric and non-parametric memories.
\newblock \emph{arXiv preprint arXiv:2212.10511}.

\bibitem[{Mu et~al.(2024)Mu, Li, and Goodman}]{mu2024learningcompresspromptsgist}
Jesse Mu, Xiang~Lisa Li, and Noah Goodman. 2024.
\newblock \href {https://arxiv.org/abs/2304.08467} {Learning to compress prompts with gist tokens}.
\newblock \emph{Preprint}, arXiv:2304.08467.

\bibitem[{Oakley and Museum(1972)}]{oakley1972man}
Kenneth~Page Oakley and London~British Museum. 1972.
\newblock \emph{Man the tool-maker}.
\newblock 538. British Museum (Natural History) London.

\bibitem[{OpenAI()}]{ChatGPT}
OpenAI.
\newblock {ChatGPT}.
\newblock \url{https://openai.com/blog/chatgpt}.

\bibitem[{Park et~al.(2023)Park, O'Brien, Cai, Morris, Liang, and Bernstein}]{park2023generativeagentsinteractivesimulacra}
Joon~Sung Park, Joseph~C. O'Brien, Carrie~J. Cai, Meredith~Ringel Morris, Percy Liang, and Michael~S. Bernstein. 2023.
\newblock \href {https://arxiv.org/abs/2304.03442} {Generative agents: Interactive simulacra of human behavior}.
\newblock \emph{Preprint}, arXiv:2304.03442.

\bibitem[{Petroni et~al.(2020)Petroni, Lewis, Piktus, Rockt{\"a}schel, Wu, Miller, and Riedel}]{petroni2020context}
Fabio Petroni, Patrick Lewis, Aleksandra Piktus, Tim Rockt{\"a}schel, Yuxiang Wu, Alexander~H Miller, and Sebastian Riedel. 2020.
\newblock How context affects language models' factual predictions.
\newblock \emph{arXiv preprint arXiv:2005.04611}.

\bibitem[{Qian et~al.(2024)Qian, Liu, Liu, Chen, Dang, Li, Yang, Chen, Su, Cong, Xu, Li, Liu, and Sun}]{qian2024chatdevcommunicativeagentssoftware}
Chen Qian, Wei Liu, Hongzhang Liu, Nuo Chen, Yufan Dang, Jiahao Li, Cheng Yang, Weize Chen, Yusheng Su, Xin Cong, Juyuan Xu, Dahai Li, Zhiyuan Liu, and Maosong Sun. 2024.
\newblock \href {https://arxiv.org/abs/2307.07924} {Chatdev: Communicative agents for software development}.
\newblock \emph{Preprint}, arXiv:2307.07924.

\bibitem[{Qiao et~al.(2024)Qiao, Zhang, Fang, Luo, Zhou, Jiang, Lv, and Chen}]{qiao2024autoact}
Shuofei Qiao, Ningyu Zhang, Runnan Fang, Yujie Luo, Wangchunshu Zhou, Yuchen~Eleanor Jiang, Chengfei Lv, and Huajun Chen. 2024.
\newblock \href {https://arxiv.org/abs/2401.05268} {Autoact: Automatic agent learning from scratch for qa via self-planning}.
\newblock \emph{Preprint}, arXiv:2401.05268.

\bibitem[{Qin et~al.(2024)Qin, Liang, Ye, Zhu, Yan, Lu, Lin, Cong, Tang, Qian, Zhao, Hong, Tian, Xie, Zhou, Gerstein, dahai li, Liu, and Sun}]{qin2024toolllm}
Yujia Qin, Shihao Liang, Yining Ye, Kunlun Zhu, Lan Yan, Yaxi Lu, Yankai Lin, Xin Cong, Xiangru Tang, Bill Qian, Sihan Zhao, Lauren Hong, Runchu Tian, Ruobing Xie, Jie Zhou, Mark Gerstein, dahai li, Zhiyuan Liu, and Maosong Sun. 2024.
\newblock \href {https://openreview.net/forum?id=dHng2O0Jjr} {Tool{LLM}: Facilitating large language models to master 16000+ real-world {API}s}.
\newblock In \emph{The Twelfth International Conference on Learning Representations}.

\bibitem[{Ram et~al.(2023)Ram, Levine, Dalmedigos, Muhlgay, Shashua, Leyton-Brown, and Shoham}]{ram2023context}
Ori Ram, Yoav Levine, Itay Dalmedigos, Dor Muhlgay, Amnon Shashua, Kevin Leyton-Brown, and Yoav Shoham. 2023.
\newblock In-context retrieval-augmented language models.
\newblock \emph{Transactions of the Association for Computational Linguistics}, 11:1316--1331.

\bibitem[{Shen et~al.(2023)Shen, Song, Tan, Li, Lu, and Zhuang}]{shen2023hugginggpt}
Yongliang Shen, Kaitao Song, Xu~Tan, Dongsheng Li, Weiming Lu, and Yueting Zhuang. 2023.
\newblock \href {https://arxiv.org/abs/2303.17580} {Hugginggpt: Solving ai tasks with chatgpt and its friends in hugging face}.
\newblock \emph{Preprint}, arXiv:2303.17580.

\bibitem[{Shi et~al.(2023)Shi, Chen, Misra, Scales, Dohan, Chi, Schärli, and Zhou}]{shi2023large}
Freda Shi, Xinyun Chen, Kanishka Misra, Nathan Scales, David Dohan, Ed~Chi, Nathanael Schärli, and Denny Zhou. 2023.
\newblock \href {https://arxiv.org/abs/2302.00093} {Large language models can be easily distracted by irrelevant context}.
\newblock \emph{Preprint}, arXiv:2302.00093.

\bibitem[{Shinn et~al.(2023)Shinn, Cassano, Berman, Gopinath, Narasimhan, and Yao}]{shinn2023reflexion}
Noah Shinn, Federico Cassano, Edward Berman, Ashwin Gopinath, Karthik Narasimhan, and Shunyu Yao. 2023.
\newblock \href {https://arxiv.org/abs/2303.11366} {Reflexion: Language agents with verbal reinforcement learning}.
\newblock \emph{Preprint}, arXiv:2303.11366.

\bibitem[{Song et~al.(2023)Song, Xiong, Zhu, Wu, Qian, Song, Huang, Li, Wang, Yao, Tian, and Li}]{song2023restgpt}
Yifan Song, Weimin Xiong, Dawei Zhu, Wenhao Wu, Han Qian, Mingbo Song, Hailiang Huang, Cheng Li, Ke~Wang, Rong Yao, Ye~Tian, and Sujian Li. 2023.
\newblock \href {https://arxiv.org/abs/2306.06624} {Restgpt: Connecting large language models with real-world restful apis}.
\newblock \emph{Preprint}, arXiv:2306.06624.

\bibitem[{Touvron et~al.(2023)Touvron, Lavril, Izacard, Martinet, Lachaux, Lacroix, Rozi{\`e}re, Goyal, Hambro, Azhar et~al.}]{touvron2023llama}
Hugo Touvron, Thibaut Lavril, Gautier Izacard, Xavier Martinet, Marie-Anne Lachaux, Timoth{\'e}e Lacroix, Baptiste Rozi{\`e}re, Naman Goyal, Eric Hambro, Faisal Azhar, et~al. 2023.
\newblock Llama: Open and efficient foundation language models.
\newblock \emph{arXiv preprint arXiv:2302.13971}.

\bibitem[{Wei et~al.(2023)Wei, Wang, Schuurmans, Bosma, Ichter, Xia, Chi, Le, and Zhou}]{wei2023chainofthought}
Jason Wei, Xuezhi Wang, Dale Schuurmans, Maarten Bosma, Brian Ichter, Fei Xia, Ed~Chi, Quoc Le, and Denny Zhou. 2023.
\newblock \href {https://arxiv.org/abs/2201.11903} {Chain-of-thought prompting elicits reasoning in large language models}.
\newblock \emph{Preprint}, arXiv:2201.11903.

\bibitem[{Wu et~al.(2023)Wu, Bansal, Zhang, Wu, Li, Zhu, Jiang, Zhang, Zhang, Liu, Awadallah, White, Burger, and Wang}]{wu2023autogenenablingnextgenllm}
Qingyun Wu, Gagan Bansal, Jieyu Zhang, Yiran Wu, Beibin Li, Erkang Zhu, Li~Jiang, Xiaoyun Zhang, Shaokun Zhang, Jiale Liu, Ahmed~Hassan Awadallah, Ryen~W White, Doug Burger, and Chi Wang. 2023.
\newblock \href {https://arxiv.org/abs/2308.08155} {Autogen: Enabling next-gen llm applications via multi-agent conversation}.
\newblock \emph{Preprint}, arXiv:2308.08155.

\bibitem[{Xu et~al.(2023)Xu, Peng, Lei, Mukherjee, Liu, and Xu}]{xu2023rewoo}
Binfeng Xu, Zhiyuan Peng, Bowen Lei, Subhabrata Mukherjee, Yuchen Liu, and Dongkuan Xu. 2023.
\newblock \href {https://arxiv.org/abs/2305.18323} {Rewoo: Decoupling reasoning from observations for efficient augmented language models}.
\newblock \emph{Preprint}, arXiv:2305.18323.

\bibitem[{Yang et~al.(2018)Yang, Qi, Zhang, Bengio, Cohen, Salakhutdinov, and Manning}]{yang2018hotpotqa}
Zhilin Yang, Peng Qi, Saizheng Zhang, Yoshua Bengio, William~W. Cohen, Ruslan Salakhutdinov, and Christopher~D. Manning. 2018.
\newblock {HotpotQA}: A dataset for diverse, explainable multi-hop question answering.
\newblock In \emph{Conference on Empirical Methods in Natural Language Processing ({EMNLP})}.

\bibitem[{Yao et~al.(2022)Yao, Zhao, Yu, Du, Shafran, Narasimhan, and Cao}]{yao2022react}
Shunyu Yao, Jeffrey Zhao, Dian Yu, Nan Du, Izhak Shafran, Karthik Narasimhan, and Yuan Cao. 2022.
\newblock React: Synergizing reasoning and acting in language models.
\newblock \emph{arXiv preprint arXiv:2210.03629}.

\bibitem[{Yin et~al.(2024)Yin, Brahman, Ravichander, Chandu, Chang, Choi, and Lin}]{yin2024agent}
Da~Yin, Faeze Brahman, Abhilasha Ravichander, Khyathi Chandu, Kai-Wei Chang, Yejin Choi, and Bill~Yuchen Lin. 2024.
\newblock \href {https://arxiv.org/abs/2311.05657} {Agent lumos: Unified and modular training for open-source language agents}.
\newblock \emph{Preprint}, arXiv:2311.05657.

\bibitem[{Yuan et~al.(2024)Yuan, Song, Chen, Tan, Shen, Kan, Li, and Yang}]{yuan2024easytool}
Siyu Yuan, Kaitao Song, Jiangjie Chen, Xu~Tan, Yongliang Shen, Ren Kan, Dongsheng Li, and Deqing Yang. 2024.
\newblock Easytool: Enhancing llm-based agents with concise tool instruction.
\newblock \emph{arXiv preprint arXiv:2401.06201}.

\bibitem[{Zhou et~al.(2023)Zhou, Schärli, Hou, Wei, Scales, Wang, Schuurmans, Cui, Bousquet, Le, and Chi}]{zhou2023leasttomost}
Denny Zhou, Nathanael Schärli, Le~Hou, Jason Wei, Nathan Scales, Xuezhi Wang, Dale Schuurmans, Claire Cui, Olivier Bousquet, Quoc Le, and Ed~Chi. 2023.
\newblock \href {https://arxiv.org/abs/2205.10625} {Least-to-most prompting enables complex reasoning in large language models}.
\newblock \emph{Preprint}, arXiv:2205.10625.

\end{thebibliography}




\appendix

\begin{figure*}[htb]
    \centering
    \includegraphics[width=\textwidth]{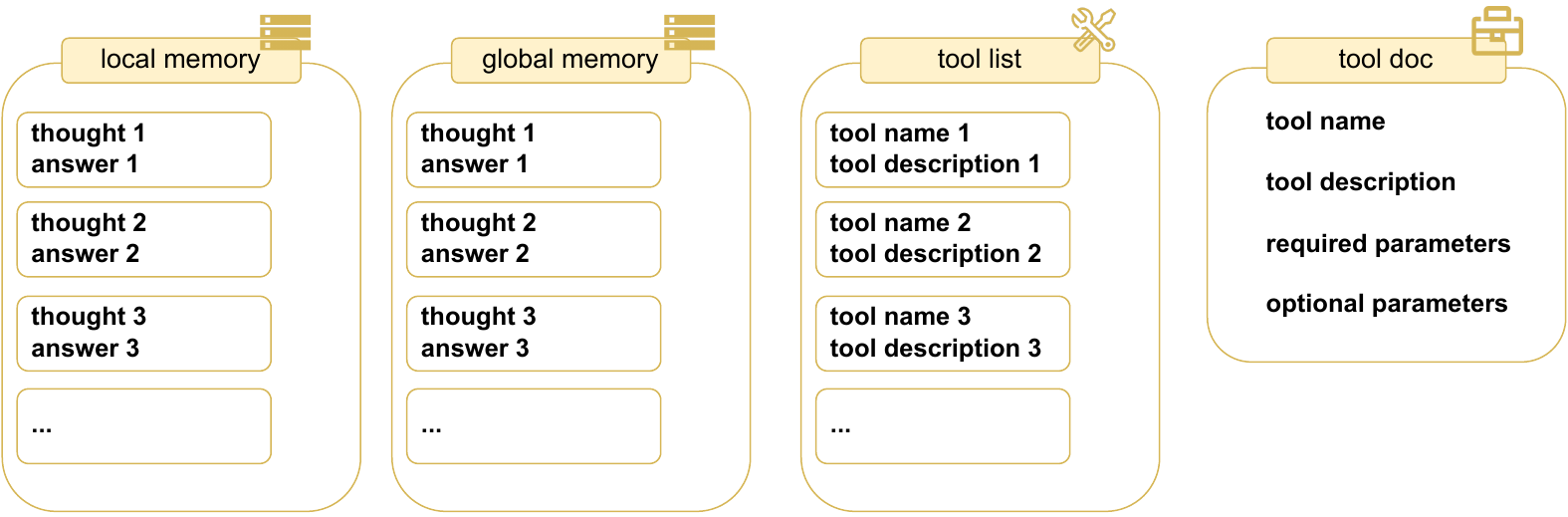}
    \caption{Demonstration of the memory of the Smurfs framework. }
    \label{fig:memory_smurfs}
\end{figure*}


\section{Comparison between Smurfs and existing MAS}
\label{appendix: Comparison}
In conclusion, Smurfs stands out compared to existing MAS in followig terms:
\paragraph{Flexibility} Smurfs utilize enhanced version of DFSDT, which intergrades global and iterative planning, while existing MAS mainly use ReAct and iterative planning. This make Smurfs more flexible in the planning process.
\paragraph{Adaptability} Smurfs realize superior performance on multiple different benchmarks, proving the adaptability of Smurfs. Existing MAS like RestGPT, Lumos are tailored for specific downstream task and need additional training to be used in other scenario.
\paragraph{Learning Efficiency} Smurfs do not need training, thus have the highest learning efficiency among existing MAS. It only need query and tool documentation from the user.

To further illustrate these points, we conducted a detailed comparison between Smurfs and two well-known Multi-Agent Systems, highlighting their differences and the adjustments required when learning out-of-box tasks.

\paragraph{CAMEL} CAMEL~\cite{li2023camelcommunicativeagentsmind} is a communicative agent framework. It uses role-play technique and inception-prompting technique to achieve autonomous cooperation between agents. CAMEL does not natively support tool use settings. CAMEL is considered to perform poorly when generalizing to new tasks (e.g., on the MATH dataset~\cite{wu2023autogenenablingnextgenllm}).

\paragraph{Autogen} Autogen~\cite{wu2023autogenenablingnextgenllm} is not designed as an agent framework for any specific task scenario. Instead, it provides a multi-agent conversation framework that allows users to customize agent characteristics and complete tasks through discussions among different agents. Autogen addresses various problems not by employing a uniform workflow, but by allowing users to design customized agents and workflows flexibly based on their tasks.

\paragraph{Smurfs} Smurfs is designed as a unified workflow specifically for complex multi-tool planning scenarios. For different tasks, Smurfs only needs to adjust the few-shot examples in the agent prompts and provide documentation for the tools applicable to the task, allowing Smurfs to generalize to other task scenarios. Smurfs was initially designed for the Stable-toolbench, which itself encompasses various types of tasks and has access to over 16,000 plugins. HotpotQA was subsequently introduced to evaluate the performance of Smurfs on closed-end tasks. When migrating from Stabletoolbench to HotpotQA, only the few-shot examples in the planning agent prompts were modified, along with the provision of plugin documentation for HotpotQA. The rest of the system continued using the same unified framework as in Stabletoolbench. Experiment results demonstrate that our untrained out-of-box unified framework achieves and even surpasses  performance of agent systems specifically trained on HotpotQA, such as Autoact and Fireact, showcasing Smurfs' flexibility in generalization.

\begin{table*}[htb!]
\centering
\scalebox{0.8}{
\begin{tabular}{|p{3cm}|p{4cm}|p{4cm}|p{5cm}|}
\hline
\textbf{Compression Method} & \textbf{Applicable Scenario}       & \textbf{Compressed Object}               & \textbf{Implementation}                                                      \\ \hline
Gist Tokens~\cite{mu2024learningcompresspromptsgist}                 & General scenarios                  & Frequently used system prompts           & Training LLM to compress system prompts, reducing token usage.               \\ \hline
CAMPHOR~\cite{fu2024camphorcollaborativeagentsmultiinput}                     & Tool use scenarios                 & Tool descriptions                        & Adopts a similar approach to gist tokens, compressing each tool description into a single token, thus reducing token cost. \\ \hline
Smurfs                      & Complex multi-tool planning scenarios & Input context in tool planning          & Operates on multi-tool planning workflows, compressing the context needed for each tool planning process. \\ \hline
\end{tabular}
}
\caption{Comparison of Token Compression Methods}
\label{tab:token compression}
\end{table*}

\section{Details of DFSDT}
\label{appendix: DFSDT Details}
DFSDT~\cite{qin2024toolllm} gives control to the model to stop and rollback the solution trajectory by using Finish tool, thus addressing limitations of ReACT. Finish tool has two parameters \textbf{give answer}: model thinks the task is finished and decide to give answer and \textbf{give up and restart}: model thinks current trajectory can't lead to correct answer and decide to rollback.
\section{Details of the Smurfs}
\label{appendix: details smurfs}
\begin{figure*}[htb]
    \centering
    \includegraphics[width=1.0\textwidth]{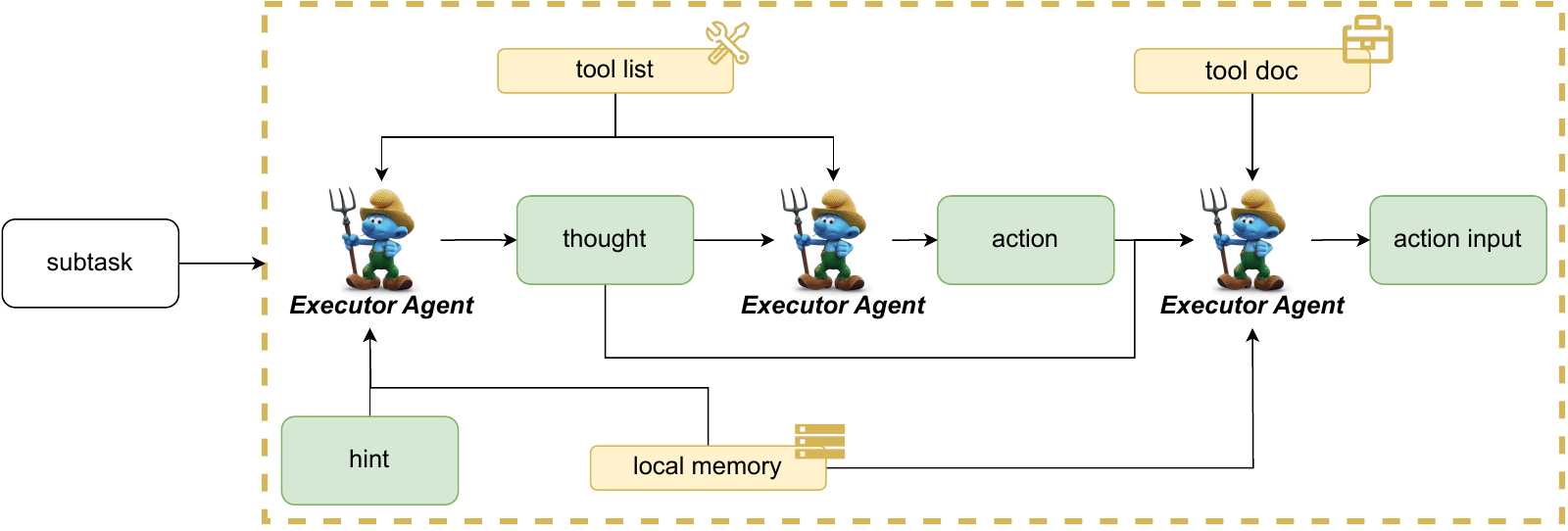}
    \caption{Details of the executor agent working process}
    \label{fig:multi-agent_detail3}
\end{figure*}

\paragraph{Executor Agent Details} As illustrated in Figure~\ref{fig:multi-agent_detail3}, given a sub-problem p, Executor Agent first thinks about what to do this time, generates thought $\gamma$ according to p, local memory M, hint h from the Verifier Agent and tool list at the current step $\tau$. Then it will choose action $\alpha$ to complete the sub-problem using p, $\gamma$ and $\tau$. After that, parameters of $\alpha$ are generated using p, local memory M and tool description of the action $D[\alpha]$. Tool is then invoked to complete the task.

\paragraph{Memory Details} As illustrated in Figure~\ref{fig:memory_smurfs}, there are four kinds of memory in Smurfs. Local memory stores thought-answer pairs of the current solution trajectory, while global memory stores all history solution trajectory (including those that is backtracked). Tool list only stores available tools' name and its usage description, while tool doc stores all detailed information about the tools including parameters details. Through using different kinds of memory under different circumstances, Smurfs can use DFSDT in a context efficient way.
\paragraph{Restart Mechanism} Every time Smurfs generate an intermediate output, a format checker is used to check whether the output is of the expected format. If not, Smurfs will retry the same step until reach retry limit or generate correct format output. This mechanism is used in addition to the rollback mechanism to handle the situation where the system can generate correct content but fail to follow the output format.

\section{Experiment Settings for Hotpot QA}
\label{appendix: experiment settings}
Following settings in \cite{qiao2024autoact}, which is randomly select 300 dev questions divided into three levels for evaluation, with 100 questions in each level. For tool library that can be used in HotpotQA, see Table~\ref{tab:tool_library}
\begin{table*}[htb]
    \centering
    \renewcommand\arraystretch{1}
    \scalebox{1.}{

    \begin{tabular}{cp{5cm}p{5cm}}
    \toprule
    \textbf{Name} & \textbf{Definition} & \textbf{Usage} \\
    \Xhline{1pt}
    BingSearch &  BingSearch engine can search for rich knowledge on the internet based on keywords, which can compensate for knowledge fallacy and knowledge outdated. &  BingSearch[query], which searches the exact detailed query on the Internet and returns the relevant information to the query. Be specific and precise with your query to increase the chances of getting relevant results. For example, Bingsearch[popular dog breeds in the United States] \\
    \hline
    Retrieve & Retrieve additional background knowledge crucial for tackling complex problems. It is especially beneficial for specialized domains like science and mathematics, providing context for the task & Retrieve[entity], which retrieves the exact entity on Wikipedia and returns the first paragraph if it exists. If not, it will return some similar entities to retrieve. For example, Retrieve[Milhouse] \\
    \hline
    Lookup & A Lookup Tool returns the next sentence containing the target string in the page from the search tool, simulating Ctrl+F functionality on the browser. & Lookup[keyword], which returns the next sentence containing the keyword in the last passage successfully found by Retrieve or BingSearch. For example, Lookup[river]. \\
    \bottomrule
    \end{tabular}
    }
    \caption{Tool library for HotpotQA.}
    \label{tab:tool_library}
\end{table*}

\section{Error Analysis on HotpotQA}
\label{appendix: Error}
To provide further insight into the model performance in the tool planning process, we manually categorized the types of error made by Smurfs in the hotpotQA hard dataset. Table~\ref{tab:Error Smurfs} shows that the most frequent errors committed by mistral-7b Smurfs is tool argument fail, followed by bad planning and answer miss. Smurfs do not make tool choice errors and premature termination errors. This shows that Smurfs actually alleviates the premature termination problem, making the tool choice process more robust. Additionally, we note that a portion of the error samples are false negatives which arise when the generated answers differ in expression from the ground truth but are equivalent in meaning. This highlights potential directions for future improvements in Smurfs.
\begin{table*}[htb!]
\centering
\scalebox{0.65}{
\begin{tabular}{@{}lccccccc@{}}
\toprule
\textbf{}                  & \textbf{Bad Planning} & \textbf{Answer Miss} & \textbf{Tool Wrong} & \textbf{Tool Argument Fail} & \textbf{False Negative} & \textbf{Premature Termination} & \textbf{Total Accuracy} \\ \midrule
\textbf{Mistral-7b-Smurfs} & 11                    & 7                     & 0                   & 28                          & 12                       & 0                              & 0.42                     \\ \bottomrule
\end{tabular}
}
\caption{Error analysis for Smurfs on HotpotQA Hard.}
\label{tab:Error Smurfs}
\end{table*}

\section{Prompts for multi-agent implementation}
\label{appendix: prompts}

Prompts used by each agent and their example outputs are shown in Figure ~\ref{fig: plan} to ~\ref{fig: verify}.

\section{Token Cost on StableToolBench Evaluation}
\label{appendix: token cost}

We analyzed the token cost for the StableToolBench experiments. As shown in Table~\ref{tab:full_token_cost}, the total token cost for each subtask within the StableToolBench is compared across three candidate tool-planning methods. The data demonstrates that, across all tasks from easy to hard, DFSDT consistently incurs high token costs, while the other two methods maintain relatively low token costs. This verifies the context-efficiency of the proposed method.

\begin{table*}
\centering
\scriptsize
\scalebox{0.77}{
    \begin{tabular}{lc|cc|cc|cc|cc|cc|cc|cc}
    \toprule
    \multirow{3}{*}{\textbf{Backbone}} & \multirow{3}{*}{\textbf{Method}} & \multicolumn{14}{c}{\textbf{StableToolBench}} \\
    & & \multicolumn{2}{c}{\textbf{I1-Inst.}} & \multicolumn{2}{c}{\textbf{I1-Cat.}} & \multicolumn{2}{c}{\textbf{I1-Tool.}} & \multicolumn{2}{c}{\textbf{I2-Cat.}} & \multicolumn{2}{c}{\textbf{I2-Inst.}} & \multicolumn{2}{c}{\textbf{I3-Inst.}} & \multicolumn{2}{c}{\textbf{Average}} \\
    & & Total & Avg. & Total & Avg. & Total & Avg. & Total & Avg. & Total & Avg. & Total & Avg. & Total & Avg. \\
    \midrule
    GPT-3.5 Turbo & ReACT & 1,010,304 & 6,198 & 824,676 & 5,390 & 1,010,514 & 6,396 & 900,855 & 7,265 & 824,510 & 7,778 & 461,121 & 7,559 & 838,663 & 6,764 \\
    GPT-3.5 Turbo & DFSDT & 3,303,062 & 20,264 & 2,745,667 & 17,945 & 3,152,532 & 19,953 & 2,560,297 & 20,648 & 3,098,365 & 29,230 & 1,390,787 & 22,800 & 2,708,452 & 21,807 \\
    \rowcolor{c5!50} GPT-3.5 Turbo & Smurfs & 1,090,404 & 7,127 & 1,917,348 & 11,763 & 1,464,535 & 9,269 & 957,088 & 7,638 & 1,096,162 & 10,341 & 632,084 & 10,362 & 1,191,270 & 9,417 \\
    \bottomrule
    \end{tabular}
}
\caption{Token costs for various candidate tool-planning methods on the StableToolBench benchmark~\cite{guo2024stabletoolbench}. `Total' indicates the total number of tokens used to complete each subtask, including both prompt and completion tokens. `Avg.' represents the average number of tokens used per question within the subtasks. Higher token counts imply greater costs for solving the same task.}
\label{tab:full_token_cost}
\end{table*}

\begin{figure*}[htb]
\begin{AIbox}{Planning Agent}
{\bf Prompt:} \\
{\footnotesize
You need to decompose a complex user's question into some simple sub-tasks and let the model execute it step by step.
Please note that: \\
1. You should only decompose this complex user's question into some simple sub-tasks which can be executed easily by using a single tool. \\
2. Each simple subtask should be expressed into natural language. \\
3. Each subtask should contain the necessary information from the original question and should be complete, explicit and self-consistent. \\
4. You must ONLY output in a parsible JSON format. An example output looks like:\\
'''\\
\{"Tasks": ["Task 1", "Task 2", ...]\}\\
'''\\
\\
This is the user's question: I'm planning a trip to Turkey and need information about postal codes in Istanbul. Can you provide me with the postal code and district for Istanbul province with plate number 34? Additionally, I would like to know if there are any transit agencies available in Istanbul. Please fetch their names and contact numbers.\\
Output: {"Tasks": ["Find the postal codes and districts for plate number 34 in Istanbul.", "Search for transit agencies and their contact numbers in Istanbul."]} \\
\\
This is the user's question: I recently moved to a new address and I need to update my information. Can you retrieve my address details using the postal code 75094080? Additionally, I would like to know the companies that offer shipping services.\\
Output: \{"Tasks": ["retrieve the address details using the postal code 75094080", "search for companies that offer shipping services to my address"]\} \\
\\
This is the user's question: \{question\}\\

Output: 
}
\tcbline
{\bf Example Output:} \\
{\footnotesize
 \{"Tasks": ["Determine the postal code and district for Istanbul province with plate number 34.", "Find out if there are any transit agencies in Istanbul.", "Get the names of the transit agencies in Istanbul.", "Obtain the contact numbers for the transit agencies in Istanbul."] \}
}
\end{AIbox} 
\caption{An example prompt for task decomposition in the framework.}
\label{fig: plan}
\end{figure*}

\begin{figure*}[htb]
\begin{AIbox}{Tool Check}
{\bf Prompt:} \\
{\footnotesize
As a powerful language model, you're equipped to answer user's question with accumulated knowledge. \\
However, in some cases, you need to use external APIs to answer accurately. \\
Thus, you need to check whether the user's question requires you to call an external API to solve it. \\
Here are some tips to help you check:  \\
1. If the user's question requires real-time information, since your knowledge base isn't updated in real-time, any such question will demand an API call. \\
2. If you need to obtain information (e.g., ID, name, phone number, geographical location, rank, etc.), you need to call the database APIs if you are not sure. \\
3. If the question demand a database search or internet research to generate an answer, this is another situation where an API call is necessary. \\
If need, please output 'YES'; If not, please output 'NO' \\
You need to give reasons first and then decide whether to keep it or not. You must only output in a parsible JSON format. Two example outputs look like: \\
Example 1: \{"Reason": "The reason why you think you do not need to call an external API to solve the user's question", "Choice": "No"\} \\
Example 2: \{"Reason": "The reason why you think you need to call an external API to solve the user's question", "Choice": "Yes"\} \\
This is the user's question: \\
\{question\}\\

Output: 
}
\tcbline
{\bf Example Output:} \\
{\footnotesize
\{"Reason": "To determine the postal code and district for a specific location based on a plate number, we would typically need to access a combination of databases, including vehicle registration databases and postal code databases. Since we do not have direct access to these databases, we will need to call external APIs to retrieve this information. Therefore, the user's question requires an API call.", "Choice": "Yes"\}
}
\end{AIbox} 
\caption{An example prompt for tool check in the framework.}
\end{figure*}

\begin{figure*}[htb]
\begin{AIbox}{Executor Agent: Thought}
{\bf Prompt:} \\
{\footnotesize
You need to analyse the previous execution history and generate your internal reasoning and thoughts on the task, and how you plan to solve it based on the current attempts.\\

Do not output thought that is too long. Output in 2-3 sentences is OK.\\

This is the user's task: \\
\{task\}\\

This is the Tool List: \\
\{functions\}\\

This is the previous execution history:\\
\{messages\}\\

This is the hint comes from the evaluator:\\
\{hint\}\\

Output:
}
\tcbline
{\bf Example Output:} \\
{\footnotesize
Looking at the task, the user wants to find the postal codes and districts for the number plate 34 in Istanbul. I have a tool that provides Turkish plates, and since Istanbul is in Turkey, this tool might provide the needed information. I will proceed to use the 'Logistics:Turkey Postal Codes:il' tool to accomplish the task.
}
\end{AIbox} 
\caption{An example prompt for tool check in the framework.}
\end{figure*}

\begin{figure*}[htb]
\begin{AIbox}{Executor Agent: Action}
{\bf Prompt:} \\
{\footnotesize
This is the user's question: \\
\{question\}\\
These are the tools you can select to solve the question:\\
Tool List:\\
\{Tool\_list\}\\

Please note that: \\
1. You should only chooce one tool from the Tool List to solve this question.\\
2. You must ONLY output the ID of the tool and your reason for choosing it in a parsible JSON format. An example output looks like:\\
'''\\
Example: \{"ID": ID of the tool, "Reason": The reason for choosing the tool\}\\
'''\\

Output: 
}
\tcbline
{\bf Example Output:} \\
{\footnotesize
\{'ID': '1', 'Reason': 'This tool provides information about Turkish plates and postal codes, which can be applied to the given task of finding the postal codes and districts for the number plate 34 in Istanbul.'\}
}
\end{AIbox} 
\caption{An example prompt for action generation in the framework.}
\end{figure*}

\begin{figure*}[htb]
\begin{AIbox}{Executor Agent: Action Input}
{\bf Prompt:} \\
{\footnotesize
Given a user's question and a API tool documentation, you need to output parameters according to the API tool documentation to successfully call the API to solve the user's question.\\
Please note that: \\
1. The Example in the API tool documentation can help you better understand the use of the API.\\
2. Ensure the parameters you output are correct. The output must contain the required parameters, and can contain the optional parameters based on the question. If no paremters in the required parameters and optional parameters, just leave it as \{\}\\
3. If the user's question mentions other APIs, you should ONLY consider the API tool documentation I give and do not consider other APIs.\\
4. The question may have dependencies on answers of other questions, so we will provide logs of previous questions and answers for your reference.\\
5. You must ONLY output in a parsible JSON Format. The example output looks like:\\
'''\\
Example: \{"keyword": "Artificial Intelligence", "language": "English"\}\\
'''\\

There are logs of previous questions and answers: \\
{previous\_log}\\

This is the current user's question: {question}\\

This is API tool documentation: {api\_dic}\\

Output:
}
\tcbline
{\bf Example Output:} \\
{\footnotesize
\{"il": 34\}
}
\end{AIbox} 
\caption{An example prompt for action input generation in the framework.}
\end{figure*}

\begin{figure*}[htb]
\begin{AIbox}{Answer Agent}
{\bf Prompt:} \\
{\footnotesize
You should answer the question based on the response output by the API tool.\\
Please note that:\\
1. Answer the question in natural language based on the API response reasonably and effectively.\\
2. The user cannot directly get API response, so you need to make full use of the response and give the information in the response that can satisfy the user's question in as much detail as possible.\\
3. Do not output answer that is too long. Output in 3-6 sentences is OK.\\

This is the user's question:\\
\{question\}\\
This is the API response:\\
\{call\_result\}\\

Output:
}
\tcbline
{\bf Example Output:} \\
{\footnotesize
The postal codes and districts for the plate number 34 in Istanbul include: Adalar district with postal code 34975 for Burgazada area, 34970 for Büyükada area, 34973 for Heybeliada area and 34977 for Kınalıada area. There is also the Arnavutköy district with postal code 34275 for areas like Anadolu, Arnavutköy Merkez, İmrahor, İslambey, Mustafa Kemal Paşa, Nenehatun, and Yavuz Selim. Another area in Arnavutköy district is Baklalı with postal code 34277.
}
\end{AIbox} 
\caption{An example prompt for Answer Agent in the framework.}
\end{figure*}

\begin{figure*}[htb]
\begin{AIbox}{Verifier Agent}
{\bf Prompt:} \\
{\footnotesize
An agent is trying to solve the query proposed by the user.\\
You need to evaluate whether the given query has been completed reasonably and accurately. If so, summarize the solution to the user. If not, summarize the current progress, and propose what is missing.\\

You response contains following elements:\\
Speak: (your words to the agent if the task is pending, or a complete answer based on the full execution log to the user if the task is finished)\\
Status: (0 or 1. 0 for pending and 1 for finished)\\

You must only output in a parsible JSON format. Two example outputs look like:\\
Example 1: \{"Speak": "answer based on the full execution log to the user", "Status": "1"\}\\
Example 2: \{"Speak": "your words to the group if the task is pending", "Status": "0"\}\\

This is the answer from the previous execution result:\\
\{answer\}\\

This is the original question: \{question\}\\

Output: 
}
\tcbline
{\bf Example Output:} \\
{\footnotesize
\{"Speak": "The postal codes and districts for plate number 34 in Istanbul include: Adalar district with postal code 34975 for Burgazada area, 34970 for Büyükada area, 34973 for Heybeliada area and 34977 for Kınalıada area. In addition, Arnavutköy district has postal code 34275 for areas such as Anadolu, Arnavutköy Merkez, İmrahor, İslambey, Mustafa Kemal Paşa, Nenehatun, and Yavuz Selim. Another part of Arnavutköy district, Baklalı, has the postal code 34277.", "Status": "1"\}
}
\end{AIbox} 
\caption{An example prompt for Verifier Agent in the framework.}
\label{fig: verify}
\end{figure*}

\end{document}